# 词义分布的空间维度——从文本符号到词向量表征[*]

厦门大学 陆晓蕾　　深湃信息科技(深圳)有限公司 倪斌

**提要**：过去几年，自然语言处理（NLP）技术飞速发展，文本表征成了计算语言学的核心。其中，分布式词向量表征在语义表达方面展现出巨大的潜力与应用效果。通过将符号化的词表征成空间向量，词向量可以解析和捕捉文本语义。本文从词向量的语言学理论基础出发，聚焦词的两种表征形式——离散式与分布式表示，简要介绍了两种表示的优缺点以及如何通过神经网络获取词向量。此外，本文介绍了词向量在语义变迁等历史语言学领域的应用。在此基础上，本文指出基于词向量的词语语义计算方法在表达多义词时存在局限性，并探讨了两种词义消歧方法：无监督与基于知识库。最后，本文提出大规模知识库与词向量的结合可能是未来文本表征研究的重要方向之一。
**关键词**：自然语言处理；文本表征；词向量

## Low-dimensional Semantic Space：from Text to Word Embedding

*LU Xiaolei*　　Xiamen University

*NI Bin*　　DeepAI Tech Co., Ltd.

**Abstract**: This article focuses on the study of Word Embedding, a feature-learning technique in Natural Language Processing that maps words or phrases to low-dimensional vectors. Beginning with the linguistic theories concerning contextual similarities — "Distributional Hypothesis" and "Context of Situation", this article introduces two ways of numerical representation of text: One-hot and Distributed Representation. In addition, this article presents statistical-based Language Models(such as Co-occurrence Matrix and Singular Value Decomposition) as well as Neural Network Language Models (NNLM, such as Continuous Bag-of-Words and Skip-Gram). This article also analyzes how Word Embedding can be applied to the study of word-sense disambiguation and diachronic linguistics.

## 1. 引言

随着人工智能与大数据研究的兴起，自然语言处理（NLP）作为一门集语言学、计算机科学于一体的跨学科研究获得了学术界和工业界的广泛关注。目前，自然语言处理主要应用

于以下四个场景：序列标注（分词/词性标注/命名实体识别/语义标注）、分类任务（文本分类/情感分析）、句子关系判断（问答系统/自然语言推理）和文本生成（机器翻译/文本摘要）。

自然语言处理的前提是文本表示（Representation），即如何将人类符号化的文本转换成计算机所能"理解"的表征形式。早期的自然语言表征主要采用离散表示。近年来，随着深度学习的不断发展，基于神经网络的分布式词向量技术在对海量语料进行算法训练的基础上，将符号化的句词嵌入到低维的稠密向量空间中，在解析句法与分析语义等方面都显示出强大的潜力与应用效果。

分布式词向量表征的核心思路是通过大量的上下文语料与算法学习，使得计算机能够自动构建上下文与目标词之间的映射关系。其主要思想是词与上下文信息可以单独构成一个可行的语义向量，这种假设具有深刻的语言学理论根源。泽利格·哈里斯（Zellig S. Harris，1954）提出分布假说（Distributional Hypothesis），认为分布相似的词，其语义也相似，这成为早期词向量表征的理论渊源之一。伦敦学派奠基人弗斯（John Rupert Firth，1957）继承并发扬了人类学家布罗尼斯拉夫·马林诺夫斯基（Bronislaw Malinowski）的"情景语境"（Context of Situation）理论，提出语境对词义的重要作用，为词向量的分布式表示与语义计算提供了思想基础。在分布假说与情景理论的基础上，词向量通过神经网络对上下文，以及上下文和目标词之间的关系进行语言建模，自动抽取特征，从而表达相对复杂的语义关系并进行语义计算。

## 2. 词的表征

作为表达语义的基本单位之一，词是自然语言处理的主要对象。进行词向量运算的前提是要将人类符号化的词进行数值或向量化表征。目前的词表征方式主要有离散式和分布式两种。

### 2.1 离散表示(One-hot Representation)

传统的基于规则的统计方法通常将词用离散的方式表示。这种方法把每个词表示为一个长向量[①]，这个向量的维度由词表[②]大小确定，并且该向量中只有一个维度的值为1，其余维度的值都为0。例如，一个语料库 A 中有三个文本，如下：

文本 1： never trouble trouble until trouble troubles you.
文本 2： trouble never sleeps.
文本 3： trouble is a friend.

那么，该语料库的词表便由[never, trouble, until, you, sleep, is, a, friend]八个单词组成。每个单词可以分别表示成一个维度为八的向量，根据单词在词表中所处的位置来计算，具体如下：{"never": [1 0 0 0 0 0 0 0]}、{"trouble": [0 1 0 0 0 0 0 0]}、…、{"a": [0 0 0 0 0 0 1 0]}、{"friend": [0 0 0 0 0 0 0 1]}。可以发现，随着语料库的变大，词表也随之增大，每个词维度也会不断变大，每个词都将成为被大量0所包围的1。因此，这种稀疏的表达方式又被形象地称为独热表示。离散表示相互独立地表示每个词，忽略了词与词在句子中的相关性，这与传统统计语言学中的朴素贝叶斯假设[③]不谋而合。然而，越来越多的实践表明，离散表达存在两大缺陷。首先是"语义鸿沟"现象，由于独热表示假定词的意义和语法是互相独立的，这种独立性显然是不适合词汇语义的比较运算，也不符合基本的语言学常识，因此，整篇文本

---

[①] 这里的长向量是维度较大的向量。在数学中，向量指具有大小和方向的量。它可以形象化地表示为带箭头的线段，空间数学可表达为[数值 1，数值 2，…，数值 n]。
[②] 语料库中的所有词构成一个词表。
[③] 朴素贝叶斯假设文本属性之间是相互独立的。

中容易出现语义断层现象。例如我们知道"端午节"与"粽子"是有联系的——端午节通常应该吃粽子。但是这两个词对应的离散向量是正交的，其余弦相关度为 0，表明两者在相似度上没有任何关系；其次是"维度灾难"，随着词表规模的增加（视语料大小，一般会达到十万以上），词向量的维度也会随之变大，向量中的 0 也会越来越多，这种维度的激增会使得数据过于稀疏，计算量陡增，并对计算机的硬件和运算能力提出更高的要求。

### 2.2 分布式表示（Distributed Representation）

为解决离散表示的两大局限性，机器需要通过分布式表示来获得低维度、具有语义表达能力的词向量（Hinton, 1986; Bengio et al. 2003）。分布式表示一般有两种方法：基于统计学和基于神经网络（详见后文三）。早期，分布式词向量的获取主要通过统计学算法，包括共现矩阵、奇异值分解等。近年来，随着深度学习技术的不断成熟，神经网络开始被用于训练分布式词向量，取代了早期的统计方法。目前分布式词向量通常特指基于神经网络获取的低维度词向量。这种词向量表示的理论源于哈里斯分布假设（Harris, 1954）：上下文相似的词，其语义也相似。分布式表示通过统计或神经网络的方法构建语言模型并获取词向量，具体方法为利用词和上下文的关系，通过算法将原本离散式的词向量嵌入到一个低纬度的连续向量空间中，最终把词表达成一个固定长度[④]的短向量。因此，这种表示方法也被称为词嵌入（Word Embedding）。此外，根据分布假设——出现在类似上下文中的单词具有类似的语义，词嵌入利用上下文与目标词的联合训练，可以获取词语的某种语义表达。例如，通过 Python 程序引入 Word2Vec 包并加载训练好的 60 维词向量模型，获得的词嵌入的形式如下："never" [1.6839292, 0.14593178, …, 0.5776881]。

| In[1]: | from gensim.models import Word2Vec    # 引入 Word2Vec 包<br>mode = Word2Vec.load("word60.model")    # 加载训练好的 60 维词向量模型<br>mode["never"]    # 获取"never"的词向量 |
|---|---|
| Out [1]: | array([1.68392922, 0.14593178, 0.44034427, 0.4281191, 1.6589873, 0.01425397, 0.54513973, 1.890714, 0.90735716, 0.30359578, 0.9542158, 0.254566, 0.23664032, 0.38192427, 0.736261, 0.02645085, 1.4055752, 0.8785525, 0.16644345, 0.7946343, 0.45621648, 0.7660677, 0.42392883, 0.51185286, 0.5855882, 1.0752455, 0.08676399, 0.5457565, 0.86771816, 0.41552883, 0.67723626, 0.28839967, 0.25073352, 1.2973395, 0.43846515, 0.23498188, 0.85107833, 1.3198441, 0.62758523, 0.06587039, 1.2524378, 1.2623415, 1.0909196, 0.17066798, 0.3085652, 0.93776286, 0.15485032, 1.2270437, 0.8903342, 0.7489391, 1.5921905, 1.6493455, 0.92749435, 0.31172457, 1.2698405, 0.79156345, 0.5858581, 1.0012968, 0.6039971, 0.5776881], dtype=float32) |

*词嵌入结果基于 Li 等（2018）的 Word2Vec 预训练词向量

## 3. 词向量训练与语言模型

目前的词表示很少采用离散表示，一般采用分布式表示。分布式词向量的获取方式可分为两种：基于统计的方法和基于神经网络的方法。

### 3.1 基于统计的方法
#### 3.1.1 共现矩阵

---

[④] 一般为 60/150/300 维。

表 1 共现矩阵单词统计结果表（上下文窗口长度为3）

|  | never | trouble | until | you | sleep | is | a | friend |
|---|---|---|---|---|---|---|---|---|
| **never** | 0 | 2 | 0 | 0 | 1 | 0 | 0 | 0 |
| **trouble** | 2 | 4 | 2 | 1 | 0 | 1 | 0 | 0 |
| **until** | 0 | 2 | 0 | 0 | 0 | 0 | 0 | 0 |
| **you** | 0 | 1 | 0 | 0 | 0 | 0 | 0 | 0 |
| **sleep** | 1 | 0 | 0 | 0 | 0 | 0 | 0 | 0 |
| **is** | 0 | 1 | 0 | 0 | 0 | 0 | 1 | 0 |
| **a** | 0 | 0 | 0 | 0 | 0 | 1 | 0 | 1 |
| **friend** | 0 | 0 | 0 | 0 | 0 | 0 | 1 | 0 |

表 1 是语料库 A 的共现矩阵单词统计结果，"never"与"trouble"共同出现的频次为 2，与"until"共同出现的频次为 0；这样，经过统计语料库 A 中的所有文本单词，"never"的词向量可以表示为[0 2 0 0 0 0 0 0]，以此类推，"trouble"可以表示为[2 4 2 1 0 1 0 0]。我们可以发现，基于词频统计结果的共现矩阵没有忽视语义关系，这在一定程度上缓和了"语义鸿沟"的问题，但是由于共现矩阵的维数等于词表的词汇总数，因此，矩阵依然十分稀疏，"维度灾难"和计算量大的问题仍然存在。

3.1.2 奇异值分解(Singular Value Decomposition, SVD)

共现矩阵的"维度灾难"与数据稀疏等问题，可以通过降低向量维度来解决，即，通过算法将共现矩阵降成低维度的稠密（dense）矩阵。奇异值分解是目前使用最为广泛的一种矩阵分解方法，可以将多维的复杂矩阵 M 分解成矩阵 U、Σ、$V^T$ 的乘积，如 $M=U\Sigma V^T$。根据奇异值的大小截取矩阵 U 后获取 U'作为降维矩阵，再经过归一化后得到词语的词向量。共现矩阵经过奇异值分解后变为低维度的稠密矩阵，该矩阵可使得语义相近的词在向量空间上相近，有时甚至能够反映词与词之间的线性关系。然而，奇异值分解算法基于简单的矩阵变换，可解释性不强；同时，由于截断操作，向量表示可能会丢失一些重要信息；另外，奇异值分解算法的计算量随语料库与词典的增长而急剧扩展，新加入的词会导致统计结果发生变化，矩阵必须重新统计和计算。

## 3.2 神经网络模型

传统的词向量模型主要基于统计学，由于近年来人工智能的快速发展，基于神经网络的语言模型愈加成熟。Xu 和 Alex（2000）最早利用神经网络构建二元（bigram[5]）语言模型的方法训练词向量。Bengio 等（2001，2003）提出了一种三层神经网络语言模型。该模型不需要人工标注语料，主要根据上文语境预测下一个词，能够从单语语料库中自动获取丰富的语义信息。该模型的提出为著名的 Word2Vec 的诞生提供了坚实的算法理论基础。

Word2Vec 是一款开源词向量工具包（Mikolov et al. 2013a），该工具包在算法理论上参考了 Benjio(2003)设计的神经网络模型，在处理大规模、超大规模的语料时，可以简单并且高效地获取高精度的词向量，在学术界和业界都获得了广泛的关注。Word2vec 的实现主要有连续词袋模型（CBOW，Continuous Bag-of-Words）和跳跃元语法模型（Skip-Gram）两种算法（图 1）。

---

[5] Gram：粒度、元。N-gram 表示多元，是计算机语言学和概率论领域内的概念，是指给定的一段文本中多个连续单位的序列。N 可以是任意正整数，如 Unigram（N=1），Bigram（N=2），Trigram（N=3），以此类推。

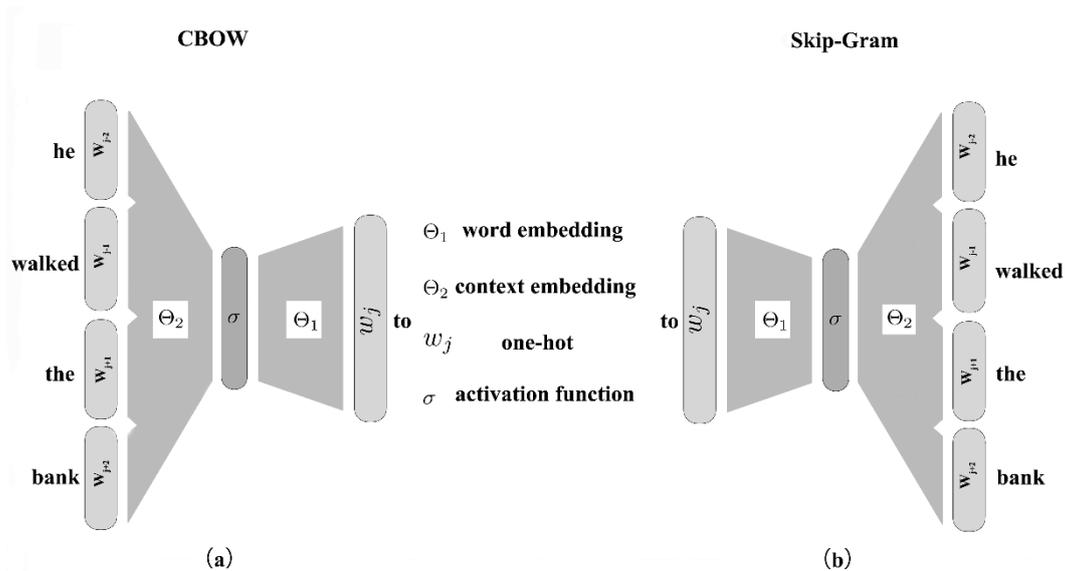

图 1 (a)连续词袋模型（CBOW）与(b)跳跃元语法模型（Skip-Gram）概念图

3.2.1 连续词袋模型（CBOW，Continuous Bag-of-Words）

连续词袋模型的核心思想是利用目标词的上下文来预测目标词出现的概率。该模型主要通过将文本视为一个词集合来训练语言模型，在运算过程中，主要考虑目标词周围出现的单词，忽略其词序和语法。因其思路类似将文字装入袋子中，这种模型也被称为词袋模型（Harris，1954）。连续词袋模型运算/运行的具体步骤为：将目标词的上下文若干个词对应的离散词向量输入模型，输出词表中所有词出现的概率，再通过哈夫曼树⑥查找目标词并通过 BP 算法⑦更新网络参数使输出为目标词的概率最大化，最终将神经网络中的参数作为目标词的词向量。如图 1a 所示，输入为"he"、"walked"、"the"、"bank"四个词的离散词向量，输出为目标词"to"的词向量。为了使得输出为"to"的概率最大，连续词袋模型通过 BP 算法不断更新神经网络参数 $\Theta_1$ 和 $\Theta_2$。经过多次迭代运算后，模型最终收敛并将运算参数（$\Theta_1$）作为单词"to"的理想词向量。

3.2.2 跳跃元语法模型（Skip-Gram）

跳跃元语法模型和词袋模型的思路相反：利用特定词语来预测其上下文。该模型接受指定的离散词向量，输出该词所对应的上下文词向量，并且通过 BP 算法更新网络参数。如图 1b 所示，输入为特定词"to"的离散向量，输出为其上下文"he"、"walked"、"the"、"bank"四个词的离散向量。同样，为了实现模型输出这四个词的（即目标词的上下文）概率最大化，Skip-Gram 通过 BP 算法更新 $\Theta_1$ 和 $\Theta_2$，并在多次迭代运算后，模型最终收敛并将获得运算参数（$\Theta_1$）作为单词"to"的理想词向量。

## 4. 语义计算与语义消歧

基于词的分布式表征以及连续词袋模型/跳跃元语法模型等神经网络模型得出的词向量，可以用于语义计算和语义消歧。传统语义计算和语义消歧主要采用语法结构分析和人工标注等消歧方法，过程复杂，人工量大。词向量技术主要通过计算机自主学习来达到消歧目的，大幅度减少了人工的投入。

### 4.1 语义（相关度）计算

---

⑥ 哈夫曼树，又称为最优树，是一种数据压缩与查找算法。
⑦ BP（Back Propagation）算法译为反向传播算法，通过结果误差的反向传播来更新神经网络参数，是深度学习的核心算法。

语义计算，即词语间的距离计算，主要用于反映语义相关度。语料经过神经网络模型运算向量化后，构成了可计算的多维向量空间。每个词在该空间内都可以表示为多维度的向量。语义计算主要的方法有两种：①通过语义词典（如著名的 WordNet 和 HowNet 等），把有关词语的概念或意义组织在一个基于语义的树形结构中，通过计算其节点（词）间的距离来反映语义的远近；②通过提取词语上下文信息，运用统计的方法进行自动计算。基于词向量空间模型的语义计算属于后者。其中，利用神经网络训练的词向量技术将文本表示为低维空间向量，通过计算向量夹角（如余弦相似度）的方式来获取词语的语义相关度。相似度取值一般为 0~1。

表 2 是基于跳跃元语法模型获取的与"语言学"相近的词，通过引入 Word2Vec 包，加载预训练的 60 维词向量模型，获得的结果按照语义相关度大小排序如下：

表 2 基于词向量计算出来的语义相关度

| from gensim.models import Word2Vec | # 引入 Word2Vec 包 |
|---|---|
| mode = Word2Vec.load("word60.model") | # 加载训练好的 60 维词向量模型 |
| mode. most_similar ("语言学") | # 获取与"语言学"相关度高的词 |
| 词汇 | 相关度（0~1） |
| 语言文学 | 0.7331274747848511 |
| 语义学 | 0.7309782505035410 |
| 语用学 | 0.7293801307678223 |
| 语音学 | 0.7213563323020935 |
| 语法学 | 0.7048327326774597 |
| 文体学 | 0.7019374966621399 |
| 词汇学 | 0.6951244473457336 |
| 翻译学 | 0.6944329142570496 |
| …… | |
| 竺可桢 | 0.3720016777515411 |
| 分配律 | 0.3720000088214874 |

*语义相关度结果基于 Li 等（2018）的 Word2Vec 预训练词向量

通过跳跃元语法模型训练出词向量后，通过类似聚类的相关度计算，可以快速（毫秒级）获取与指定词汇语义相关的词汇。结果显示，"语言文学"、"语义学"等与"语言学"的相关度较高，在空间位置上较为接近；"竺可桢"、"分配律"等与"语言学"的相关度较低。这类语义计算对于语义聚类以及语义挖掘有一定的价值。值得注意的是，分布式词向量技术对语料的依赖程度较高，因此，需要精选语料进行大规模学习以实现偏差最小化。

通过计算向量相关度，基于词向量的自然语言处理技术能够从海量的语料中快速获取词语语义的相对位置，并查找出与之相似的词。多维的词向量经过降维后，可以在二维平面上清晰地看出语义关系。例如，图 2 中，在词汇关系类比中，"king"与"man"之间的距离与"queen"与"woman"之间的相对位置非常接近。在句法类比中，"slow-slower-slowest"三词之间的相对距离与"fast-faster-fastest"和"long-longer-longest"等的相对位置也十分相似。

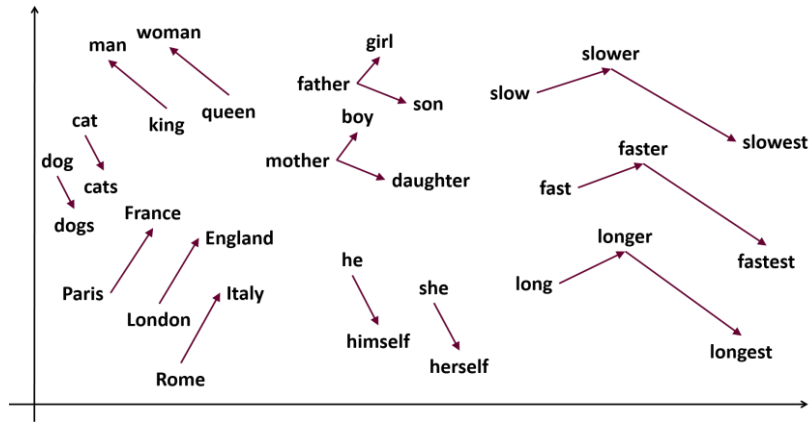

图 2 降维后的词向量空间（Akhtar, 2018）

以上，可以发现词向量在语义相关度计算与句法分析上可以做到定量分析与可视化，这对语义挖掘具有十分重要的应用价值。

## 4.2 语义消歧

词向量技术虽然可以表征语义，然而在面对多义词的时候，单个向量很难表达词语的多个意义，依然存在词义模糊以及多义消失（Meaning Conflation）等问题。因此，在使用词向量时，需要考虑歧义对结果的影响。传统的语义消歧主要通过语法结构（Reifler，1955），建立特定领域的语义库以减少语义数（Weaver，1955），通过人工标注的语料学习消歧规则建立词汇专家系统（Weiss,1973）等，大多依赖人工建立的语义网络与语义角色。在深度学习领域，消歧主要根据目标词的上下文信息来进行。目前，语义消歧方法基本上可以分为两类：无监督式（Ranganato, et al. 2017; Chang & Sui, 2018）和基于知识库（Tripodi, 2017; Chaplot & Salakhutdinov, 2018）的方式。无监督的方式直接从文本语料中学习意义，而基于知识库的方式则在计算机深度学习的基础上，利用人类专家制作的外部语言知识库作为意义来源，将机器学习与专家知识相结合。前者可解释性较差，后者融合专家归纳整理的知识库，解释性较好，但也因受限于知识库，对知识库以外的词汇和意义泛化性不足。

### 4.2.1 无监督消歧

无监督的方式主要有语境聚类式（Liu, et al.，2015）、混合式（Li, et al.，2015）和语篇主题嵌入式（Huang, et al.，2012）等方法。语境聚类式（Clustering-based）消歧的主要思想是通过收集单词出现的语境，利用聚类算法对其词义进行自动分类。混合式（Joint-training）消歧主要通过在训练的过程中加入词义比对更新模式，自动生成词义组来实现。语篇主题嵌入式主要通过在局部信息（Local Context）的基础上引入全局信息(Global Context)来实现消歧。相对而言，语篇主题嵌入的方法能够获得更为精准的语义消歧效果。

在词向量训练过程中，一般不考虑整个篇章，仅利用句子上下文几个窗口的词提供的信息来训练模型。然而，有些具有歧义的词义无法仅凭单句上下文几个词的信息来判断。如图 3 的英文句子"he walks to the ***bank***"中，bank 可能可以被理解为"银行"或者"河岸"。此时，语篇主题嵌入式消歧会在词向量训练中加入全局信息和多种词义原型（Multiple Word Prototypes），具体如下：

第一，全局信息模型将整个篇章的词向量做加权平均（Weighted Average，权重是 tf-idf）计算后作为全局语义向量（Global Semantic Vector），再和正常训练的局部语义向量相加，这样训练出来的加强型词向量能更好地捕捉语义信息。例如，篇章里出现的诸如"river"、"play"、"shore"、"water"等词，可以使得当前"bank"的语义为"河岸"的概率大大增强。

第二，使用多个词向量代表多义词。通过对上下文的词向量进行加权平均（代表目标词

语义）后进行 K 均值聚类，根据聚类结果作为目标词的意义类别，如 bank$^1$、bank$^2$ 和 bank$^3$。显然，这种方式将词根据语义的不同来分别训练词向量，在某种程度上突破了多义消失的问题。然而，调查发现这种方法的效果强烈依赖于聚类算法的可靠性，也不可避免地存在误差。

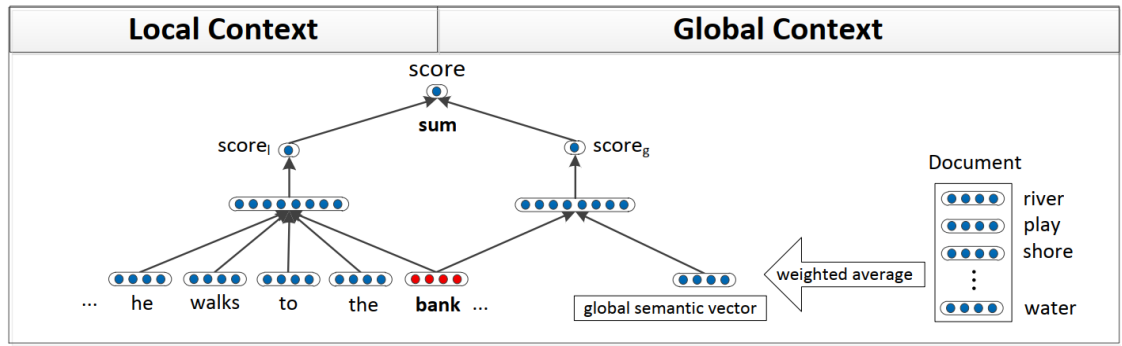

图 3 语篇主题嵌入式（Huang et al.，2012）

4.2.2 基于知识的方法

所谓基于知识的方法，即在词向量的训练过程中，加入其它结构化的知识作为监督。随着以 WordNet 与 HowNet 为代表的语言知识库的不断完善，基于其网络结构的图模型方法也逐渐用于语义消歧中。监督学习借助有标注的训练语料，在特定领域已经获得了较好的消歧性能（Agirre et al.，2009）。

Yu 等（2014）在训练连续词袋模型的同时，引入 PPDB 数据库⑧和 WordNet 等外部知识，抽取语义相似词对作为约束条件，使得对应的词向量能够学习到这些词义相似的信息。Bian 等（2014）在连续词袋模型中加入词的形态、句法和语义信息。Nguyen 等（2016）在跳跃元语法模型基础上加入词汇对比信息共同训练，使得训练得到的词向量能有效识别同义词和反义词。Niu 等（2017）将 HowNet 知识融入到词向量连续词袋模型与跳跃元语法模型中，训练词义的最基本粒度——义原（Sememe）⑨，在训练过程中加入上下文—单词—意义—义原的联合训练，有效地提升词向量在表达多义词的效果。

以上，无监督消歧单纯依靠语料挖掘意义，极大的减少了人工的投入，而基于知识的方式则引入了外部语料知识，有效地克服了因缺乏足够信息导致的语义不完整等困难。

## 5. 词向量与语义变迁

词汇作为语言的基本单位，其语义变迁是研究语言模型和反映社会历史文化演变的重要手段。传统的语义变迁研究主要通过从历史文本中搜索目标词，统计词汇的使用频次，根据语言和历史知识对其进行人工描述。Michel 等（2011）利用 Google Books 五百多万种出版物，建立语料库，通过词频统计研究人类文化的演变与特点。Bamman 等（2011）则通过观测与目标词汇共现的其他词汇的频度变化来间接地探索词汇语义变化。Mihalcea 等（2012）通过收集 19 世纪至 21 世纪特定术语的使用变化来考察社会现象。以上工作大多通过搜索和统计的方法，从海量的文献中捕捉到了各个历史时期的词汇语义，费时费力，且难以直观获取语义内涵。而词向量表征将文本转换为空间向量，用向量的夹角代表其语义相似度，能够定量地从海量历时文本中获取语义相近的词。通过研究词汇的语义相近词，能够比较直观地看出语义的历时变化。

刘知远等（2016）基于 1950 年至 2003 年的《人民日报》文本训练词向量模型，对词汇语义变化进行了定量观测，探究了词汇变化反映出来的社会变迁。Hamilton 等（2018）在多

---

⑧ PPDB 为一种基于农药特性的专业数据库。
⑨ 义原在语言学中是指最小的不可再分的语义单位，知网（HowNet）是最著名的义原知识库。

语言大规模语料库的基础上，利用 Word2Vec 的跳跃元语法模型建立历时词向量空间来揭示语义变迁规律。如图 4a 中，"gay"在二十世纪初与"tasteful"、"cheerful"等词汇在空间位置上较为接近，到了世纪中叶，"gay"与"witty"、"bright"等词的语义相关度高。到了二十世纪末，"gay"的词义与"homosexual"与"lesbian"等词在语义计算上结果相近。图 4b 显示，随着报纸、电视、广播、网络等多种媒体的兴起，"broadcast"的相似词也从十九世纪中期的"seed"、"sow"等，逐渐演变为"newspapers"、"television"、"radio"、"bbc"等。图 4c 揭示了"awful"的语义从十九世纪中期的"solemn"逐渐向"terrible"、"appalling"等演变的过程。Hamilton 等（2018）通过动态建模，将静态的词向量扩展到动态的时间序列场景中，定量地观测与剖析了语义更迭与社会文化的变迁。

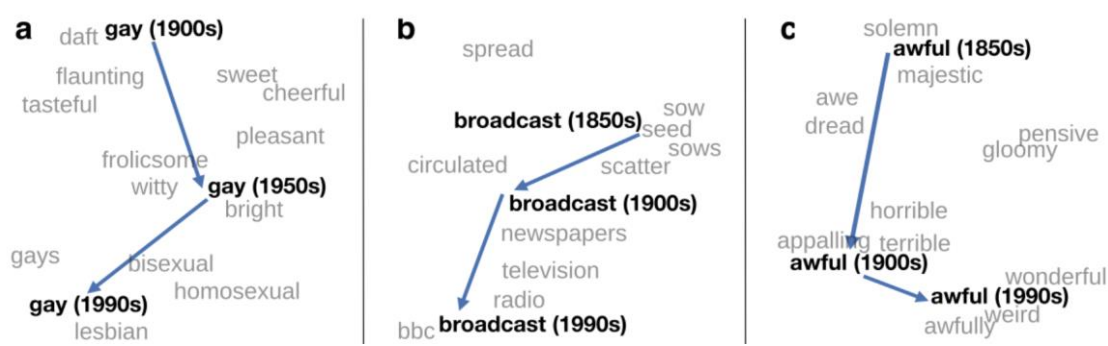

图 4 "gay"、"broadcast"、"awful"等词的语义变迁示意图（Hamilton et al.，2018）

## 6. 结语

本文深入探讨了词向量在表达语义方面的表现，介绍了两种词向量表达的形式以及获取方式。证明了词向量技术为语义消歧与语义变迁等研究提供了定量手段，在语义表达方面显示出强大的潜力与应用效果。

分布式词向量模型是基于海量语料的监督学习，充分利用语料库中词的上下文相关信息，通过神经网络优化训练语言模型，在此过程中获得词语的向量化形式。这种向量化的分布式表征以"情景语境"为理论基础，通过向量间的夹角余弦相似度来度量词汇的相似度。但是，我们也发现现阶段的词向量仅仅从海量的语料库中学习到部分语义表达，在其歧义性和不常用词的弱表达上尚且不尽人意，单从海量数据中学到的语义表达还是存在偏差。另外，词向量对于训练语料库中未出现的词也很难去表达其语义。针对这种情况，本文认为在文本以外，应该引入更加强大的人类专家知识库的支持，获取更加强大的语义表达。为此，词向量的研究，乃至整个自然语言处理系统需要探索数据与知识共同驱动的方法，不断完善语义表征算法，扩充与优化语言专家知识体系。

**References [参考文献]**

Agirre E, De Lacalle O L, Soroa A. 2009. Knowledge-based WSD on Specific Domains: Performing Better than Generic Supervised WSD. Twenty-First International Joint Conference on Artificial Intelligence. IJCAI.

Akhtar, S. S. 2018. *Robust Representation Learning for Low Resource Languages*. INDIA: International Institute of Information Technology.

Bamman D, Crane G. 2011. Measuring Historical Word Sense Variation. Proceedings of the 11th Annual International ACM/IEEE Joint Conference on Digital Libraries. ACM.

Bengio Y, Ducharme R, Vincent P, et al. 2003. A Neural Probabilistic Language Model. *Journal of*

*Machine Learning Research* 3: 1137-1155.

Bian J, Gao B, Liu T Y. 2014. Knowledge-powered Deep Learning for Word Embedding. Joint European Conference on Machine Learning and Knowledge Discovery in Databases. Springer.

Firth J R. 1957. A Synopsis of Linguistic Theory, 1930-1955. *Studies in Linguistic Analysis*.

Hamilton W L, Leskovec J, Jurafsky D. 2016. Diachronic Word Embeddings Reveal Statistical Laws of Semantic Change[J]. *arXiv preprint arXiv*:1605.09096.

Harris Z S. 1954. Distributional Structure[J]. *Word*, 10(2-3): 146-162.

Hinton G E. 1986. Learning Distributed Representations of Concepts. Proceedings of the Eighth Annual Conference of the Cognitive Science Society. CSS 1: 12.

Huang E H, Socher R, Manning C D, et al. 2012. Improving Word Representations Via Global Context and Multiple Word Prototypes. Proceedings of the 50th Annual Meeting of the Association for Computational Linguistics: Long Papers-Volume 1. Association for Computational Linguistics 873-882.

Li J, Jurafsky D. 2015. Do Multi-sense Embeddings Improve Natural Language Understanding?. *arXiv preprint arXiv*:1506.01070.

Li S, Zhao Z, Hu R, et al. 2018. Analogical Reasoning on Chinese Morphological and Semantic Relations. *arXiv preprint arXiv*:1805.06504.

Liu P, Qiu X, Huang X. 2015. Learning Context-sensitive Word Embeddings with Neural Tensor Skip-gram Model. Twenty-Fourth International Joint Conference on Artificial Intelligence. IJCAI.

Michel J B, Shen Y K, Aiden A P, et al. 2011. Quantitative Analysis of Culture Using Millions of Digitized Books. *Science* 331(6014): 176-182.

Mihalcea R, Nastase V. 2012. Word Epoch Disambiguation: Finding How Words Change Over Time. Proceedings of the 50th Annual Meeting of the Association for Computational Linguistics. ACL.

Mikolov T, Chen K, Corrado G, et al. 2013b. Efficient Estimation of Word Representations in Vector Space. *arXiv preprint arXiv*:1301.3781.

Mikolov T, Sutskever I, Chen K, et al. 2013a. Distributed Representations of Words and Phrases and Their Compositionality. Advances in Neural Information Processing Systems. NIPS.

Miller G A. 1995. WordNet: a Lexical Database for English. *Communications of the ACM* 38(11): 39-41.

Nguyen K A, Walde S S, Vu N T. 2016. Integrating Distributional Lexical Contrast into Word Embeddings for Antonym-synonym Distinction[J]. *arXiv preprint arXiv*:1605.07766.

Niu Y, Xie R, Liu Z, et al. 2017. Improved Word Representation Learning with Sememes[A]. Proceedings of the 55th Annual Meeting of the Association for Computational Linguistics. ACL.

Reifler E. 1955. The Mechanical Determination of Meaning[J]. *Readings in Machine Translation*: 21-36.

Weaver Warren. 1955. Translation[J]. *Machine Translation of Languages* 14: 15-23.

Weiss S F. 1973. Learning to disambiguate[J]. *Information Storage and Retrieval* 9(1): 33-41.

Xu W, Rudnicky A. 2000. Can Artificial Neural Networks Learn Language Models?. Sixth International Conference on Spoken Language Processing. ISCA.

Yu M, Dredze M. 2014. Improving Lexical Embeddings with Semantic Knowledge. Proceedings of the 52nd Annual Meeting of the Association for Computational Linguistics. ACL.

Liu, Zhiyuan(刘知远), Liu Yang(刘扬), Tu Cunchao(涂存超), Sun Maosong(孙茂松).2016. Lexical Semantic Variation and Social Change: Quantitative Observation and Analysis. *Chinese Journal of Language Policy and Planning* 1(6): 47-54. [2016, 词汇语义变化与社会变迁定量观测与分析. 《语言战略研究》第 1 期: 47-54.